\pdfoutput=1

\documentclass[11pt]{article}

\usepackage{naacl2021}

\usepackage{times}
\usepackage{latexsym}

\usepackage[T1]{fontenc}

\usepackage[utf8]{inputenc}

\usepackage{amsmath}
\usepackage{amssymb}
\usepackage{color,soul}

\usepackage{microtype}
\usepackage{graphicx}
\usepackage{subfigure}
\usepackage{booktabs} 
\usepackage{multirow}
\usepackage{amsmath}
\usepackage{adjustbox}
\usepackage{lipsum}
\usepackage{xcolor}
\usepackage{microtype}

\title{Supervised Relation Classification as Two-way Span-Prediction}

\author{Amir DN Cohen \\
  Bar Ilan University \\
  \texttt{amirdnc@gmail} \\\And
  Shachar Rosenman \\
  Bar Ilan University \\
  \texttt{shacharosn@gmail}\\\And
  Yoav Goldberg \\
  Bar Ilan University \& AI2\\
  \texttt{yoav.goldberg@gmail.com} \\}

\date{}

\begin{document}
\maketitle
\begin{abstract}
The current supervised relation classification (RC) task uses a single embedding to represent the relation between a pair of entities. We argue that a better approach is to treat the RC task as span-prediction (SP) problem, similar to Question answering (QA). We present a span-prediction based system for RC and evaluate its performance compared to the embedding based system. We demonstrate that the supervised SP objective works significantly better then the standard classification based objective. We achieve state-of-the-art results on the TACRED and SemEval task 8 datasets.
\end{abstract}
\section{Introduction}
\begin{figure*}[h]
\centering
\includegraphics[width=1\textwidth]{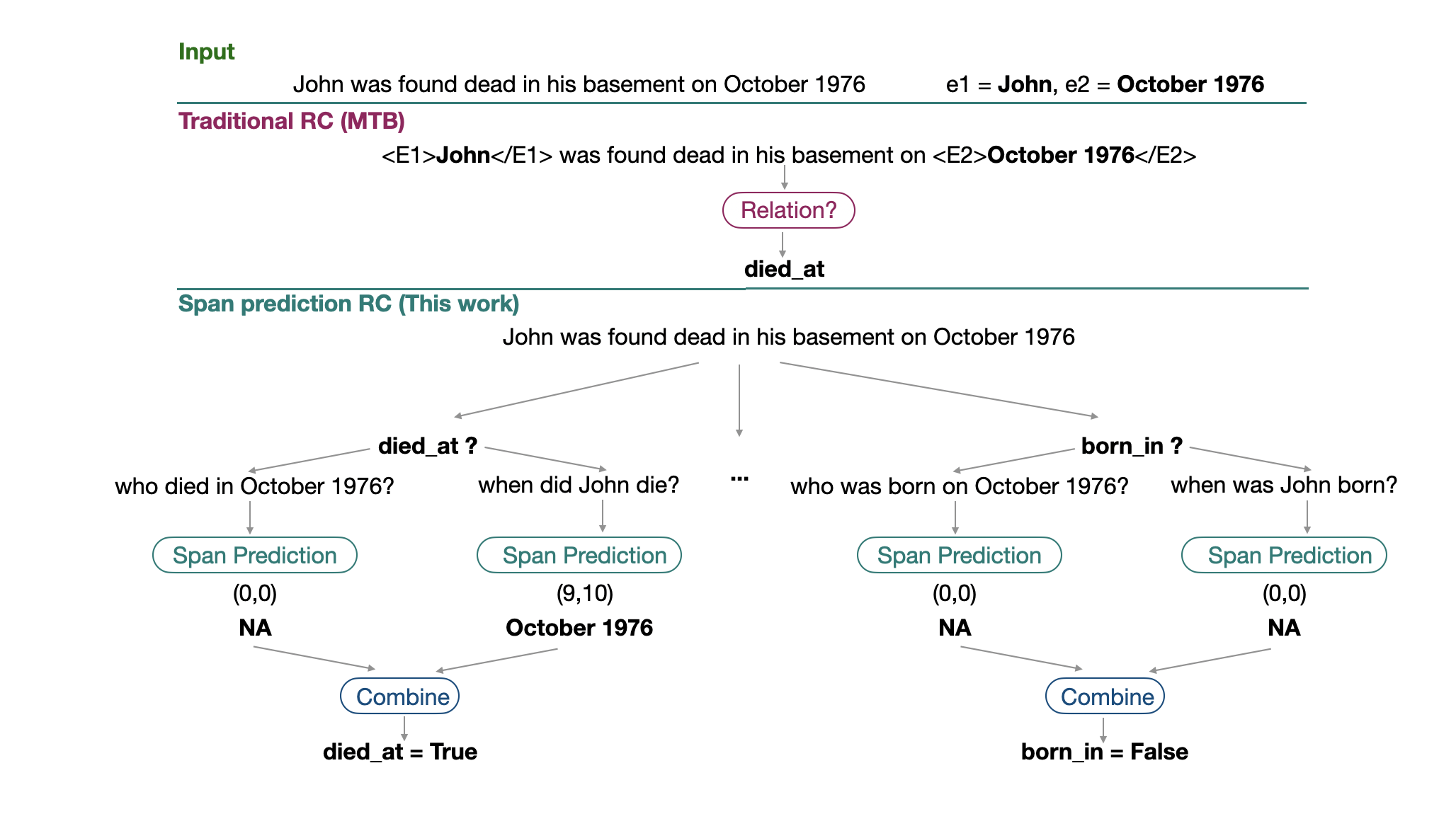}
\caption{Traditional RC (top) VS our span-prediction approach (bottom). for each relation type that is compatible with the marked entity type, we create two questions. If the model answers one of them correctly, we assert the relation over the two entities.} \label{fig:main}
\end{figure*}

The relation classification (RC) task revolves around binary relations (such as "[$e_1$] founded [$e_2$]") that hold between two entities. The task is to read a corpus and return entity pairs $e_1, e_2$ for which the relation holds (according to the text). This is often posed as a Relation Classification task (RC), in which we are given a sentence and two entities (where each entity is a span over the sentence), and need to classify the relation into one of $|R|$ possible relations, or to a null ``no-relation" class if none of the $|R|$ relations hold between the given entities. Relation Extraction datasets, including the popular and large TACRED dataset \cite{Zhang}, all take the relation classification view, by providing tuples of the form $(s, e_1, e_2, r)$, where $s$ is a sentence, $e_1, e_2$ are entities in $s$ and $r$ is a semantic relation between $e_1$ and $e_2$. Consequently, all state-of-the-art models follow the classification view: the sentence and entities are encoded into a vector representation, which is then being classified into one of the $R$ relations. The training objective then aims to embed the sentence + entities into a space in which the different relations are well separated.
We argue that this is a sub-optimal training architecture and training objective for the task, and propose to use span-predictions models, as used in question-answering models, as an alternative.
Our method converts RC datasets to the SP form, using this reduction we evaluate the RC using SP models in a supervised setting. We show a high level flow of our method in Figure \ref{fig:main}.

\citet{Alt2020} analyzed the errors of current RC systems and showed that 10\% of the errors are the result of predicting a relation that is based on other arguments in the sentence. This was further explored by \cite{Rosenman2020}, where the authors showed that the current embedding based methods often classify a sentence without considering the marked entities. Our span-prediction method forces the model to identify the exact entities which compose each relation, which helps the model to overcome the challenges presented in these two works. 
We demonstrate this on the TACRED and SemEval datasets. Our method surpasses the current state-of-the-art on these datasets by $2.3 F_1$ points on TACRED and $0.9 F_1$ points on SemEval. Additionally, we experiment with the newly ``released challenge relation extraction'' (CRE) dataset, which was made specifically to test the existence of shallow heuristics in RE models. On all three datasets, our span-prediction models outperform existing RC. We also experiment with several different templates and show that our method can benefit from templates that add some prior semantic knowledge that is related to the classified relation type. All the method we present and our experiments will are publicly available online.

\section{Related Work}
Using QA for \emph{zero-shot RE/RC} has been explored by \citet{Levy2017},
who demonstrated that by posing each relation as a question, one can train a system that transfers well to new and unseen relations in a ``zero-shot'' setting. We show that by improving their reduction by making it two-directional and switching to a fully supervised setup we can improve the accuracy of state-of-the-art RC system. Following \citet{Levy2017}, several works proposed the use of 
 SP like architectures to solve a variety of tasks like coreference resolution \cite{Wu2019}, event extractions \cite{Du2020}, nested named entities \cite{Li2019b} and multi turn entity extraction \cite{Li2019}.
While the mentioned works used SP models to improve performance on a specific task, It's worth mentioning that other works have used QA for different reasons, like \cite{He2015} that used QA as an easier way to annotate data for the SRL task.

Recently, \citet{Jiang2019} presented a unified model for many NLP tasks, using a unified span-based classification method. Such predictors may also benefit from adopting a QA-like span modeling, as we present here.







\section{Embedding Classification vs Span-Prediction}
\paragraph{Embedding-Based Relation Classification} \label{ssec:rc}
A \emph{RC sample} takes the form $(c, e_1, e_2, r)$ where $c = [c_0, \ldots, c_n]$ is a context (usually a sentence), $e_1$ and $e_2$ are spans that correspond to head and tail entities and are given as spans over the sentence, and $r \in R\cup\{\emptyset\}$ is a relation from a predefined set of relations $R$, or $\emptyset$ indicating that no relation from $R$ holds.
\emph{RC classifier} takes the form of a multi-class classifier:
\[f_{rc}(c, e_1, e_2) \mapsto r \in R\cup\{\emptyset\}\]
\noindent The training objective is to score the correct $r \in R\cup\{\emptyset\}$ overall incorrect answers, usually using a cross-entropy loss. State-of-the-art methods \cite{Soares2019} achieve this by learning an embedding function $embed(c, e_1, e_2)$ that maps instances with the same relation to be close to the embedding of the corresponding relation in an embedding space. The embedding function is used on pre-trained masked LMs such as SpanBERT \cite{Joshi2019}, RoBERTa \cite{Liu2019} and ALBERT \cite{Lan2019}.

\paragraph{Span Prediction}
A \emph{SP sample} takes the from of $(c, q, e_a)$ where $c = [c_0, \ldots, c_m]$ is a context (a sentence or a paragraph), $q = [q_0, \ldots, q_l]$ is a query, and $e_a$ is the answer to the query represented as a span over the $c$, or a special out-of-sequence-span indicating that the answer does not exist.\footnote{In practice, this span is the out-of-sentence \emph{CLS} token.} 

\noindent\emph{SP model} takes the form of a \emph{span predictor} from a $c,q$ pair to a span over $c$:
\[f_{qa}(c, q) \mapsto e_a \in [1 .. m] \times [1 .. m] \]
\noindent This predictor takes the form: \[\arg\max_{e_a}score_{c,q}(e_a)\] 
\noindent where $score_{c,q}(e_a)$ is a learned span scoring function, and $e_a$ ranges over all possible spans. The training objective is to maximize the score the correct spans above all other candidate spans. The scoring function in state-of-the-art models \cite{McCann2018,He2015,Wu2019} also make use of pre-trained LMs. 


\subsection{Method Comparison} \label{ssec:compare}
The question $q$ (``where was Sam born?'') in QA can be thought of as involving a span $e_q$ (``Sam'') and a predicate $r_q$ (``where was born''). Under this view, the SP classifier can be written as:
\[f_{qa}(c,e_q,r_q)\mapsto e\] 
compared to the relation classifier: 
\[f_{rc}(c,e_1,e_2) \mapsto r\] Note that both methods include a context, two spans, and a relation/predicate, but the RC models classify from two spans to a relation (from a fixed set), while the SP model classify from a span and a relation (from a potentially open set) into another span.

Let's review the implications of this difference:\\

\noindent\textbf{Embedding} 
While the two methods embeds the input prior to classification, the items that are being embedded change. In RC the embedding $h_{re}$ is based on the context and entities:
\[ h_{re} = embed(c, e_1, e_2) \]
while for span-prediction the embedding $h_{qa}$
 encodes both the context and the question (the relation of interest and one of the entities):\footnote{In practice, the embedding is obtained via a pre-trained LM such as BERT, and as per-usual is prefixed with a \texttt{CLS} token, while the different components are separated with a \texttt{SEP} token.}
\[ h_{qa} = embed(q, c) = embed(r, e_1, c) \]

Note that the span-predictor embedding includes the relation name, as well as template word that surround the $(r, e_1)$ pair. This makes the embedding strictly more informative, and has several benefits, as we explain below.

\subsection{Implications}
\paragraph{Relation type indication for the pretrained model.}
The inclusion of the relation $r$ in the input to the contextualized embedder allows the embedder to specialize on a specific relation. For example, consider the sentence ``\emph{Martha gave birth to John last February}''. The entity John participates in two relations: ``date of birth'' and ``parents of''. The RC embedding will have to either infer the relation based on the entities, or else preserve information regarding both relations, while in the span-prediction case the embedding takes the relation $r$ into account, and can focus on the existence (or nonexistence) of one of the entities as the argument for this relation. Focusing on a specific relation in the embedding stage (which involves most of the computation of the model) allows using all of the model computation for a specific relation.

\paragraph{Sharing of semantic information.}
The span-prediction model is based on templates encoding $r$ and $e$, and these templates may pass valuable information to the model:
(1) by containing semantic information that is correlated to the target relation (e.g. questions that represent the relation); and
(2) by containing information that can help generalize over different relations.

For example, consider the relation ``born in'' with the template question ``Who was born in X?'' and the relation ``parent of'' with the question ``Who is the parent of X?''. While the relations are different from each other, they both contain an entity of type ``person'', a similarity which is communicated to the model by the use of the shared word ``Who''. This can help the model generalize commonalities across relation types, when needed.

\paragraph{More demanding loss function.}
During training, relation-classification models classify sentences with marked entities to one of $|R| + 1$ relation types. Span prediction models are also required to decide whether the sentence contains a given relation (they should predict if the sentence contains the answer or not), but they are also required to predict the span of the missing argument. This means that the span-prediction models are required to predict the relation between the input entities in addition to the relation itself.

\begin{figure*}[t!]
\centering
\begin{tabular}{llll}
\hline
\textbf{RC sample}                       & \textbf{\begin{tabular}[c]{@{}l@{}}Relation \\ candidates\end{tabular}} & \textbf{\begin{tabular}[c]{@{}l@{}}Question\\ (Reverse Question)\end{tabular}} & \textbf{Answer} \\ \hline
\multirow{4}{*}{\textbf{John} was born on \textbf{1991}}   & \multirow{2}{*}{\ul{"Date of birth"}}                                        & When was John Born?                                                            & 1991            \\ \cline{4-4} 
                                         &                                                                         & (Who was born on 1991?)                                                        & John            \\ \cline{2-4} 
                                         & \multirow{2}{*}{"Date of death"}                                        & When did john die?                                                            & N/A             \\ \cline{4-4} 
                                         &                                                                         & (Who died on 1991?)                                                            & N/A             \\ \hline
\multirow{6}{*}{\textbf{Mary} is \textbf{John}'s employer} & \multirow{2}{*}{\ul{"employer of"}}                                          & Who is employed by Mary?                                                       & John            \\ \cline{4-4} 
                                         &                                                                         & (Who is John employer?)                                                        & Mary            \\ \cline{2-4} 
                                         & \multirow{2}{*}{"siblings"}                                             & Who is the sibling of Mary?                                                    & N/A             \\ \cline{4-4} 
                                         &                                                                         & (Who is the sibling of John?)                                                  & N/A             \\ \cline{2-4} 
                                         & \multirow{2}{*}{"parents of"}                                           & Who is the child of Mary?                                                      & N/A             \\ \cline{4-4} 
                                         &                                                                         & (Who is the parent of John?)                                                   & N/A             \\ \hline
\end{tabular}
\caption{\textbf{Supervised dataset construction}. Example of span-prediction samples that are generated from RC samples. The RC sample contains sentence, entities (highlighted) and relation, while the span-prediction sample has a context (same as the sentence the RC sentence), a query, and an answer. A set of relation questions are created based on the RC entities types; underlined relations are the correct ones.}
\label{fig:mrcqa-example}
\end{figure*}

\paragraph{Limitations.} It is important to note, however, that the span-prediction method is more computationally expensive: instead of performing a single contextualized embedding operation followed by $k+1$-way classification, we need to perform $k$ contextualized embedding operations (and in our case, $2k$ such operations), each of them followed by scoring of all spans. We leave ways of improving the computational efficiency of the model to future work.




\section{Reducing RC to span-prediction} \label{sec:method}
Given the uncovered similarity between RC and span-predicting showed in Section \ref{ssec:compare}, we now describe how to reduce RC to SP.

Given an RC instance $(c, e_1, e_2) \mapsto rel$ we can create an SP instance $(q=(e_q, rel_q), c) \mapsto e_a$ as follows. Let $T_{rel}(e)$ be a \emph{template function} associated with relation $rel$. The function takes an entity $e$ and returns a question. For example, a template for date-of-birth relation might be $T_{dob}=$``\textit{When was \underline{\hspace{1em}} born}'', and $T_{dob}(\text{Sam})=$``\textit{When was Sam born?}''. Given an RC instance $(c, e_1, e_2) \mapsto rel$ we can now create a span-prediction instance $(T_{rel}(e_1), c) \mapsto e_2$, and return that the relation $rel$ holds if the span returned from $f_{qa}(T_{rel}(e_1), c)$ is compatible with $e_2$. This is essentially the construction used by \citep{Levy2017}. 

\paragraph{Bidirectional questions.}
We note that the decision to predict $e_2$ based on $e_1$ is arbitrary, and that we could have just as well change the template to e.g. ``Who was born on \underline{\hspace{1em}}?'', and predict $e_1$ from $e_2$.

We propose to use both options, by associating a relation $rel$ with \emph{two} templates, $T^{e_1\rightarrow e_2}_{rel}$ and $T^{e_2\rightarrow e_1}_{rel}$, creating the two corresponding SP instances, and combining the two answers. Concretely, given the RC instance:\\[0.5em]
\indent RC:$(c, Sam, 1991) \mapsto \textit{date-of-birth}$\\[0.5em] we create the two SP instances:\\[0.5em]
\indent QA1:$(c, \textit{When was Sam born?}) \mapsto \textit{1991}$\\[0.5em]
\indent QA2:$(c, \textit{Who was born in 1991?}) \mapsto \textit{Sam}$\\[0.5em]
We show in Section \ref{sec:experements} that using two questions indeed results in substantial improvements.
\paragraph{Template formulation.}
Note that while in this example we formulate the questions in English, simpler template might also be used. We also experiment with a template that replaces the question by the relation name and another template that used an unused token for each relation. We elaborate on the template variations in detail in Section \ref{sec:experements}.
\paragraph{Answer combination.}
There are various possible strategies to combining the two answers. An approach which we found to be effective is to combine using an OR operation: if either of the returned spans is compatible with the expected span,\footnote{Two non-empty spans are said to be compatible if either of them contain the other.}  the relation $rel$ is returned, and if neither of them is compatible, the answer is no-relation.

A natural alternative is to combine using an AND operation, requiring the answers of the two questions to be compatible in order to return $rel$. In our experiments (Section \ref{sec:ablation}), this yielded lower F-scores on the relation classification task, as we classified more cases as no-relation when we shouldn't have. The span predictor network had easier time answering one formulation on some instances, and the other formulation on others. As span-prediction models quality improve, future applications may reconsider the combination method.

\paragraph{Binary vs. Multiclass.} This reduction targets a binary version of RC, where the relation is given and the classifier needs to decide if it holds or not. We extended it to the multi-class version by creating a version for each of the relevant\footnote{
A relation is relevant for a given pair of entities if the entities types match that of the relation.} relations.\footnote{In the rare case (less than 4\%) that our model predicts more than one relation, we return one of them arbitrarily.}
\paragraph{Supervised dataset construction.}
The reduction allows to train a SP model to classify RC instances. For each RC training instance $(c, e_1, e_2, r)$, where $r \in R\cup\{\emptyset\}$, we consider all relations $r' \in R$ which are compatible with $(e_1, e_2)$.\footnote{A relation is compatible with a pair of entities if it is between entities with the same named-entity types.}
We then generate two SP instances for each of the compatible relations. Instances that are generated with the templates of the gold-relation $r$ are marked as positive instances (their answer is either $e_1$ or $e_2$, as appropriate), while instances that are generated from $r' \neq r$ are negative examples (their answer is the no-answer span). Figure \ref{fig:mrcqa-example} provides an example.

\paragraph{Per-template thresholds.}
General purpose SP models use a global threshold $\tau$ to distinguish between answerable and non-answerable questions given a context. 
In the supervised relation classification case, the set of questions is fixed in advance to $2|R|$. We observe that the optimal threshold value for each question is different. We thus set a different threshold value $\tau^i_{rel}$ for each template. The threshold is set using the model’s threshold setting procedure, but considering the set of questions generated from each template separately.

\section{Main Experiments} \label{sec:experements}
\subsection{Datasets}
We compare ourselves on three RC datasets.
\paragraph{TACRED} \cite{Zhang} is the currently most popular and largest RC dataset.
It spans 41 ``classic'' RC relations, which hold between persons, locations, organizations, dates, and so on (e.g, "siblings", "dates of birth", "subsidiaries", etc).
TACRED contains 106,264 labeled sentences (train + dev + test), where $20\%$ of the data is composed from the 41 relations and the rest $80\%$ are ``no relation'' instances. 
\paragraph{SemEval 2010 Task 8} (SemEval, \citet{Hendrickx2010}), is a smaller dataset, containing 10,717 annotated examples covering 9 relations, without no-relation examples.  
SemEval relations are substantially different from those in TACRED, covering more abstract relations such as part-whole, cause-effect, content-container, and so on.
\paragraph{Challenge relation extraction (CRE)} \citet{Rosenman2020} showed that current RC models have a strong bias towards shallow heuristic that does not capture the deep semantically relation between entities. For example, classifying an entity pair by the entities type + an unrelated event in the sentence. To show this bias empirically, they created a Wikipedia based dataset intended to be used only for testing, which contains 3000 manually tagged sentences from the TACRED relations. Each sentence in the dataset contains two entity pairs that are compatible with the same relation. The evaluation of the CRE is binary - the model goal is to indicate if a given relation is found or not found in the dataset. The model was evaluated with both SP and RC models.

\subsection{Template variations.} \label{ssec:supervised}

We convert the TACRED and SemEval training sets to span-prediction form in three ways, representing various amounts of semantic information. From most informative to least, the variations are: 
\paragraph{Natural language questions (question)} For each RC sample we create two samples, as described in Section \ref{sec:method}. The complete template list is available in the supplementary materials.
\paragraph{Relation name (relation)} Same as the question dataset, but we replace each of the questions with the relation name, entity, and a marker that indicate if it's a head or tail entity. E.g., the relation\\ 
\indent RC:$(c, John, CEO) \mapsto \textit{per:title}$\\[0.5em] will be represented as the questions:\\[0.5em]
\indent QA1:$(c, \textit{per:title t John}) \mapsto \textit{CEO}$\\[0.5em]
\indent QA2:$(c, \textit{per:title h CEO}) \mapsto \textit{John}$
\paragraph{Unique tokens (token)} Same as the relation dataset, but we replace the relation name with a new reserved token. E.g., the above \emph{per:title} relation will be represented as the questions:\\[0.5em]
\indent QA1:$(c, \textit{r2 t John}) \mapsto \textit{CEO}$\\[0.5em]
\indent QA2:$(c, \textit{r2 h CEO}) \mapsto \textit{John}$\\[0.2em]

Each of the datasets used the same train/validation/test splits. 

\subsection{Comparisons}
We compare our results to several leading models, reporting the results from the corresponding papers.
\paragraph{MTB} \cite{Soares2019} is a state of the art RC model which is based on BERT-large, and which does not involve any additional training material except for the pre-trained LM. MTB way of creating sentence embedding is the current SOTA, and thus our most direct comparison.\footnote{The same paper reports additional results based on external training data, which is not comparable. However, these results have since been superseded by the KEPLER model.}

\paragraph{KEPLER} \cite{Wang2019} 
This model holds the current highest reported RC results over TACRED.
It is a RoBERTa based RC model which incorporates additional knowledge in the form of a knowledge-graph derived from Wikipedia and Wikidata and uses MTB for sentence embedding. 

\paragraph{LiTian} \cite{Li2020} is the current top-scoring model on the SemEval dataset. It uses a dedicated RC architecture and uses the BERT pre-trained LM. 

\subsection{Models}
We train span-predicting models using the architecture described in \cite{Devlin2018}, starting from either the BERT-Large \cite{Devlin2018} or ALBERT \cite{Lan2019} pre-trained LMs.\footnote{We used the implementations provided by Huggingface \cite{Wolf2019}. Following previous work, used the Adam optimizer, an initial learning rate of $3e^{-5}$, and up to 20,000 steps with early stopping on a dev-set.}

BERT-large is used to compare the SOTA model reported in \cite{Soares2019} on equal grounds, while ALBERT is a stronger pre-trained LM which is used to show the full capabilities of our approach.\footnote{We also ran preliminary tests using \cite{Liu2019} and \cite{Joshi2019} that showed inferior results compared to ALBERT.}

\section{Main Results}

\begin{table}[t]
\centering
\begin{tabular}{l c c c}
\toprule
Model  & Acc$_+$ & Acc$_-$ & Acc \\
\midrule
\midrule
RC$_{BERT}$ & 70.0 & 64.8 & 67.1 \\
SQuAD$_{BERT}$ & 62.9 & 70.9 & 67.4 \\
SP$_{token,BERT}$ & 55.0 & 75.5 & 66.4 \\
SP$_{relation,BERT}$ & 66.6 & 72.1 & 69.6 \\
SP$_{question,BERT}$ & {\bf 72.5} & {\bf 75.0} & {\bf 73.9} \\
\midrule
\midrule
SQuAD$_{ALBERT}$ & 71.5 & 78.8 & 75.3 \\
SP$_{token,ALBERT}$ & 80.9 & 73.2 & 76.6 \\
SP$_{relation,ALBERT}$ & 78.2 & 79.8 & 79.1 \\
SP$_{question,ALBERT}$ & {\bf 81.2} & {\bf 79.5} & {\bf 80.3} \\
\bottomrule
\end{tabular} 
\caption{{\bf CRE.} Span prediction model result on CRE, compared to traditional RC and QA model. RC models are relation classification models and SQuAD models are QA models that were trained on the SQuAD 2.0 dataset.}
\label{tab:challange}
\end{table}

\begin{table}[t!]
\centering
\scalebox{0.9}{
\begin{tabular}{l c c c}
\toprule
Model  & P & R & F$_1$ \\
\midrule
\midrule
RC$_{MTB,BERT}$ & - & - & 70.1 \\
SP$_{token,BERT}$ & 63.3 & 78.4 & 70.0 \\
SP$_{relation,BERT}$ & 67.0 & 76.0 & 71.2 \\
SP$_{question,BERT}$ & {\bf 71.1} & {\bf 72.6} & \textbf{71.8} \\
\midrule
\midrule
KEPLLER$_{RoBERTa + KG}$ & 72.8 & 72.2 & 72.5\\
SP$_{token,ALBERT}$ & 72.2 & 74.6 & 73.4 \\
SP$_{relation,ALBERT}$ & {\bf 74.6} & {\bf 75.2} & \textbf{74.8} \\
SP$_{question,ALBERT}$ & 73.3 & 71.8 & 72.6 \\
\bottomrule
\end{tabular}}
\caption{{\bf TACRED.} Supervised results on the  TACRED datasets. {\bf Top}: Using BERT. This is a direct comparison to the MTB span-prediction model. MTB F$_1$ is taken from the original paper. SP models (except token) suppress MTB. {\bf Bottom}: Using ALBERT. Here the reference point is KEPLLER, the current best performing model on this dataset. All the supervised SP-ALBERT models outperform KEPPLER.}
\label{tab:supervised_tacred}
\end{table}

\begin{table}[t!]
\centering
\scalebox{1.0}{
\begin{tabular}{l c c c}
\toprule
Model  & P & R & F$_1$ \\
\midrule
\midrule
RC$_{MTB,BERT}$ & - & - & 89.2 \\
LiTian (current best) & \bf{ 94.2} & 88.0 & 91.0 \\
SP$_{token,BERT}$ &  92.8 & 88.8 & 90.7 \\
SP$_{relation,BERT}$ & 91.9 & 83.1 & 87.1 \\
SP$_{question,BERT}$ & 90.7 & \bf{ 93.2} & \textbf{91.9} \\
\bottomrule
\end{tabular}}
\caption{{\bf SemEval.} Supervised results on the  SemEval datasets. LiTian is the current state of the art. }
\label{tab:supervised_semeval}
\end{table}
The results of the CRE evaluation are presented in Table \ref{tab:challange}. We report the results in the same format used in the original paper: the percentage of positive samples that were identified correctly (Acc$_+$), the percentage of negative samples that were identified correctly (Acc$_-$), and the overall weighted accuracy (Acc). Except for the token-BERT reduction, all of the reduction we used surpassed their RC and SQuAD trained models, where the SP model (BERT and ALBERT) improve by more then 5\% compared to the squad models. We also observed a correlation between the amount of semantic information in the templates and the model performance.

The results for TACRED are presented in  Table \ref{tab:supervised_tacred}, both our BERT based SP and relation datasets outperform MTB model. like in CRE, there is a clear correlation between the amount of semantic data in the template and the model accuracy. This is somewhat surprising considering the fact that the amount of semantic data given to each relation template in the QA reduction is negligible compared to the amount of data the model see during training.

The results for SemEval are presented in Table \ref{tab:supervised_semeval}. The best performing model is the QA model, which also suppress LiTian's model. Surprisingly, the token model perform better than the relation token, which somewhat undermine our hypothesis that the semantic information in the templates correlates to the model overall accuracy. We explain this anomaly by looking at the relation names in SemEval. In contract to TACRED (and CRE) the relation names in SemEval are somewhat abstract, and have lower semantic similarity to the relation instances. For example, the TACRED relation "per:parents" gives a lot more information than the SemEval relation "instrument-agency".

Another difference between the datasets is the difference in accuracy gain from each of our models CRE has the most benefit, then TACRED and then SemEval. We assume that this difference originates from the dataset nature - The "shallow huristics" shown by \cite{Rosenman2020} that CRE was made to highlight are more prominent in TACRED then SemEval. . Our span-prediction based loss is specially tailored to deal with such situations. In contrast, SemEval does not contain the "no relation" type, and the chance of any two relations to appear in the same sentence is low, resulting in this challenge to be a lot less prominent on SemEval then TACRED. 

Since we didn't have access to KEPLLER (the current SOTA), we used the best pretrained model available to us - ALBERT model. all of our ALBERT-based relation reduction methods outperforms the current best TACRED model (KEPLER) by 2.3\% $F_1$, despite KEPLER using external data. 

Another anomaly is that on ALBERT, the QA reduction performed worse by the relation reduction and even the token reduction. This somewhat undermines our assumption that QA reduction is superior to relation reduction because the former contains more information about the relation type. We currently have no convincing explanation to this result. 

\section{Ablations} \label{sec:ablation}

\begin{table}[t]
\scalebox{0.98}{
\centering
\begin{tabular}{l c c c}
\toprule
Model  & P & R & F$_1$ \\
\midrule
\midrule
Two Questions (full model)  & 73.3 & 71.8 & 72.6 \\
Single Question & 75.8 & 65.4 & 70.2 \\
\bottomrule
\end{tabular}}
\caption{{\bf Importance of bidirectional questions.} The SP$_{question,ALBERT}$ model with two questions combined via OR (full setup), vs. a single question.  Asking two questions instead of one significantly increase the model performance.}
\label{tab:ablation}
\end{table}

\begin{table}[t]
\scalebox{0.77}{
\centering
\begin{tabular}{l c c c}
\toprule
Model  & P & R & $F_1$ \\
\midrule
SP$_{token,BERT}$ & 80.1 (63.3) & 54.7 (78.4) & 65.0 (70.0) \\
SP$_{relation,BERT}$ & 84.4 (67.0) & 44.8 (76.0) & 58.5 (71.2) \\
SP$_{question,BERT}$ & 83.15 (71.1) & 50.0 (72.6) & 62.4 (71.8) \\
\midrule
\midrule
SP$_{token,ALBERT}$ & 80.1 (72.2) & 54.7 (74.6) & 65.0 (73.4) \\
SP$_{relation,ALBERT}$ & 81.9 (74.6) & 56.1 (75.2) & 66.6 (74.8) \\
SP$_{question,ALBERT}$ & 81.2 (73.3) & 55.9 (71.8) & 66.3 (72.6) \\
\bottomrule
\end{tabular} }
\caption{{\bf Answer combination method.} TACRED performance when using the AND combination (in brackets, the corresponding OR combination). Using AND substantially increase precision, while dropping recall, resulting in a lower F$_1$ score on all models.}
\label{tab:and}
\end{table}

\paragraph{The importance of bidirectional questions.} To assess the impact of using questions in both directions, we also report the ALBERT-based QA-reduction on TACRED in which we present two questions per relation, but where both questions use $e_1$ as the template argument and $e_2$ as the answer (``Single Question'' in Table \ref{tab:ablation}). This model has significantly less success than the two-way model, resulting in a drop of $2.4\% F_1$. 


\paragraph{Combination using OR vs. AND.}
We combine the answers of the two generated questions by an "OR" operator, but the same can be done with the "AND" operator. To check this we ran our models but report the relation as "present" iff the two questions return a correct answer. In Table \ref{tab:and} we report the results. The AND operator greatly under-perform when compared to the OR operator with a drop in $F_1$ of about 10\%. The reason for this degradation is that the AND operator is more focused on precision, while the OR operator is more focused on recall. Over the years a major challenge of RC system was to increase recall \cite{Ji} - It's easier for RC system to filter unrelated samples then to generalize to new patterns.

\begin{table}[t]
\scalebox{0.96}{
\centering
\begin{tabular}{l c c c}
\toprule
Model  & P & R & $F_1$ \\
\midrule
SQuAD 2.0 (Zero shot) & 49.7 & {\bf 78.9} & 57.1 \\
\midrule
SP+Pretrain (BERT,unified) & 68.3 & 63.2 & 65.5 \\
SP+Pretrain (BERT,serial) & 70.1 & 65.1 & 67.5 \\
\midrule
SP$_{question,BERT}$ & {\bf 71.1} & 72.6 & {\bf 71.8} \\
\midrule
\end{tabular}}
\caption{{\bf Using SQuAD 2.0} Top: Evaluating SQuAD 2.0 QA model on TACRED in a zero-shot setup, using our bidirectional SP reduction. Mid:
``Fine-tuning'' the SP$_{question,BERT}$ models on TACRED after SQuAD 2.0 per-training. Bottom: The SP model trained with out pre-training, significantly outperforming the pre-trained variants.}
\label{tab:unified}
\end{table}

\section{Relation to SQuAD Training}

We advocated a fully-supervised training of RC models as span-prediction. How well does this compare to using existing QA models, like SQuAD, in a zero-shot setting? And can we leverage the existing knowledge in QA datasets, via pre-training? We explore these two options, and conclude that while the zero shot accuracy is impressively high, the unification of SQuAD and TACRED harm the overall accuracy. 

\paragraph{Zero-shot SQuAD} In light of the success of SQuAD trained model on CRE (as demonstrated by \citet{Rosenman2020}), we evaluate the SQuAD 2.0 trained model performance on TACRED, using our bidirectional reduction. In this zero shot setup we take a SQuAD trained model (without any modifications) and apply our reduction to evaluate the test set of TACRED.

\paragraph{Joined training with SQuAD}
We now attempt to leverage the SQuAD 2.0 data  to improve our RC model. We train our SP$_{question,BERT}$ model by combining the data original SQuAD questions and the TACRED-generated questions. We do this in two ways: in the \textbf{unified} version we combine the two datasets simply by shuffling together the TACRED and SQuAD questions into a single dataset. In the \textbf{serial} version we first train on the SQuAD data, and then continue training the model on TACRED data.

\paragraph{Results} 
Table \ref{tab:unified} lists the results.
Unsurprisingly, the zero shots F$_1$ score on TACRED is are substantially lower than all the supervised variants. However, the recall of the zero shot setup is substantially higher: the SQuAD 2.0 model is very permissive. 

Interestingly, the additional SQuAD questions did not improve---and even substantially hurt---the SP method compared to train on only the TACRED-generated questions. This goes to highlight that the main benefits of the SP method comes from the combination of the supervised training and the span-prediction objective, and not merely from the QA form, or from the additional semantic information that is potentially embedded in the QA models.

\section{Conclusion}
In this work, we argue for the use of span-prediction methods, typically used for QA, to replace the standard RC architectures. Our approach reduce each RC sample to a series of binary span-prediction tasks. We Show that This approach achieves state-of-the-art performance in supervised settings, with the moderate cost of supplying question templates that describe the relation.

\bibliography{anthology,custom}
\bibliographystyle{acl_natbib}

\clearpage 
\appendix
\section*{Question Templates For TACRED}
Tables \ref{tbl:tmp1} and \ref{tbl:tmp2} on the next page show the question templates we used on the TACRED dataset for the QA reduction. 

\begin{table*}[h]
\scalebox{1}{
\begin{tabular}{|l|l|}
\hline
\textbf{Relation Name} & \textbf{Question} \\ \hline
\multirow{2}{*}{ per:date\_of\_birth } & Q1:  When was $e_1$ born?                 \\
               & Q2 Who was born in $e_2$?                 \\ \hline
\multirow{2}{*}{ per:title } & Q1:  What is $e_1$'s title?                 \\
               & Q2 Who has the title $e_2$                 \\ \hline
\multirow{2}{*}{ org:top\_members/employees } & Q1:  Who are the top members of the organization $e_1$?                 \\
               & Q2 What organization is $e_2$ a top member of?                 \\ \hline
\multirow{2}{*}{ org:country\_of\_headquarters } & Q1:  In what country the headquarters of $e_1$ is?                 \\
               & Q2 What organization have it's headquarters in $e_2$?                 \\ \hline
\multirow{2}{*}{ per:parents } & Q1:  Who are the parents of $e_1$?                 \\
               & Q2 Who are the children of $e_2$?                 \\ \hline
\multirow{2}{*}{ per:age } & Q1:  What is $e_1$'s age?                 \\
               & Q2 Whose age is $e_2$?                 \\ \hline
\multirow{2}{*}{ per:countries\_of\_residence } & Q1:  What country does $e_1$ resides in?                 \\
               & Q2 Who resides in country $e_2$?                 \\ \hline
\multirow{2}{*}{ per:children } & Q1:  Who are the children of $e_1$?                 \\
               & Q2 Who are the parents of $e_2$?                 \\ \hline
\multirow{2}{*}{ org:alternate\_names } & Q1:  What is the alternative name of the organization $e_1$?                 \\
               & Q2 What is the alternative name of the organization $e_2$?                 \\ \hline
\multirow{2}{*}{ per:charges } & Q1:  What are the charges of $e_1$?                 \\
               & Q2 Who was charged in $e_2$?                 \\ \hline
\multirow{2}{*}{ per:cities\_of\_residence } & Q1:  What city does $e_1$ resides in?                 \\
               & Q2 Who resides in city $e_2$?                 \\ \hline
\multirow{2}{*}{ per:origin } & Q1:  What is $e_1$ origin?                 \\
               & Q2 Who originates from $e_2$?                 \\ \hline
\multirow{2}{*}{ org:founded\_by } & Q1:  Who founded $e_1$?                 \\
               & Q2 What did $e_2$ found?                 \\ \hline
\multirow{2}{*}{ per:employee\_of } & Q1:  Where does $e_1$ work?                 \\
               & Q2 Who is an employee of $e_2$?                 \\ \hline
\multirow{2}{*}{ per:siblings } & Q1:  Who is the sibling of $e_1$?                 \\
               & Q2 Who is the sibling of $e_2$?                 \\ \hline
\multirow{2}{*}{ per:alternate\_names } & Q1:  What is the alternative name of $e_1$?                 \\
               & Q2 What is the alternative name of $e_2$?                 \\ \hline
\multirow{2}{*}{ org:website } & Q1:  What is the URL of $e_1$?                 \\
               & Q2 What organization have the URL $e_2$?                 \\ \hline
\multirow{2}{*}{ per:religion } & Q1:  What is the religion of $e_1$                 \\
               & Q2 Who believe in $e_2$                 \\ \hline
\multirow{2}{*}{ per:stateorprovince\_of\_death } & Q1:  Where did $e_1$ died?                 \\
               & Q2 Who died in $e_2$?                 \\ \hline
\multirow{2}{*}{ org:parents } & Q1:  What organization is the parent organization of $e_1$?                 \\
               & Q2 What organization is the child organization of $e_2$?                 \\ \hline
\multirow{2}{*}{ org:subsidiaries } & Q1:  What organization is the child organization of $e_1$?                 \\
               & Q2 What organization is the parent organization of $e_2$?                 \\ \hline
\end{tabular}}
\caption{TACRED question templates part 1}
\label{tbl:tmp1}
\end{table*}

\begin{table*}[]
\scalebox{0.9}{
\begin{tabular}{|l|l|}
\hline
\textbf{Relation Name} & \textbf{Question} \\ \hline
\multirow{2}{*}{ per:other\_family } & Q1:  Who are family of $e_1$?                 \\
               & Q2 Who are family of $e_2$?                 \\ \hline
\multirow{2}{*}{ per:stateorprovinces\_of\_residence } & Q1:  What is the state of residence of $e_1$?                 \\
               & Q2 Who lives in the state of $e_2$?                 \\ \hline
\multirow{2}{*}{ org:members } & Q1:  Who is a member of the organization $e_1$?                 \\
               & Q2 What organization $e_2$ is member of?                 \\ \hline
\multirow{2}{*}{ per:cause\_of\_death } & Q1:  How did $e_1$ died?                 \\
               & Q2 How died by $e_2$?                 \\ \hline
\multirow{2}{*}{ org:member\_of } & Q1:  What is the group the organization $e_1$ is member of?                 \\
               & Q2 What organization is a member of $e_2$?                 \\ \hline
\multirow{2}{*}{ org:number\_of\_employees/members } & Q1:  How many members does $e_1$ have?                 \\
               & Q2 What organization have $e_2$ members?                 \\ \hline
\multirow{2}{*}{ per:country\_of\_birth } & Q1:  In what country was $e_1$ born                 \\
               & Q2 Who was born in the country $e_2$?                 \\ \hline
\multirow{2}{*}{ org:shareholders } & Q1:  Who hold shares of $e_1$?                 \\
               & Q2 What organization does $e_2$ have shares of?                 \\ \hline
\multirow{2}{*}{ org:stateorprovince\_of\_headquarters } & Q1:  What is the state or province of the headquarters of $e_1$?                 \\
               & Q2 What organization's headquarters are in the state or province $e_2$?                 \\ \hline
\multirow{2}{*}{ per:city\_of\_death } & Q1:  In what city did $e_1$ died?                 \\
               & Q2 Who died in the city $e_2$?                 \\ \hline
\multirow{2}{*}{ per:city\_of\_birth } & Q1:  In what city was $e_1$ born?                 \\
               & Q2 Who was born in the city $e_2$?                 \\ \hline
\multirow{2}{*}{ per:spouse } & Q1:  Who is the spouse of $e_1$?                 \\
               & Q2 Who is the spouse of $e_2$?                 \\ \hline
\multirow{2}{*}{ org:city\_of\_headquarters } & Q1:  Where are the headquarters of $e_1$?                 \\
               & Q2 Which organization has its headquarters in $e_2$?                 \\ \hline
\multirow{2}{*}{ per:date\_of\_death } & Q1:  When did $e_1$ die?                 \\
               & Q2 Who died on $e_2$                 \\ \hline
\multirow{2}{*}{ per:schools\_attended } & Q1:  Which schools did $e_1$ attend?                 \\
               & Q2 Who attended $e_2$?                 \\ \hline
\multirow{2}{*}{ org:political/religious\_affiliation } & Q1:  What is $e_1$ political or religious affiliation?                 \\
               & Q2 Which organization has is political or religious affiliation with $e_2$?                 \\ \hline
\multirow{2}{*}{ per:country\_of\_death } & Q1:  Where did $e_1$ die?                 \\
               & Q2 Who dies in $e_2$?                 \\ \hline
\multirow{2}{*}{ org:founded } & Q1:  When was $e_1$ founded?                 \\
               & Q2 What organization was founded on $e_2$?                 \\ \hline
\multirow{2}{*}{ per:stateorprovince\_of\_birth } & Q1:  In what state was $e_1$ born?                 \\
               & Q2 Who was born in state $e_2$?                 \\ \hline
\multirow{2}{*}{ org:dissolved } & Q1:  When was $e_1$ dissolved?                 \\
               & Q2 Which organization was dissolved in $e_2$?                 \\ \hline
\end{tabular}}
\caption{TACRED question templates part 2}
\label{tbl:tmp2}
\end{table*}

\end{document}